\documentclass{article}
\usepackage{spconf,amsmath,graphicx}

\usepackage{enumitem}
\setlist{nosep, leftmargin=14pt}

\usepackage{mwe} % to get dummy images

\usepackage{graphicx}
\usepackage{color}
\usepackage{amssymb}
\usepackage{booktabs}
\usepackage{lipsum}
\newcommand\blfootnote[1]{%
\begingroup
\renewcommand\thefootnote{}\footnote{#1}%
\addtocounter{footnote}{-1}%
\endgroup
}

\title{Learning Shape Priors by Pairwise Comparison for Robust Semantic Segmentation}
\name{Cong Xie\textsuperscript{1,2}, Hualuo Liu\textsuperscript{2,3}, Shilei Cao\textsuperscript{2}, Dong Wei\textsuperscript{2}, Kai Ma\textsuperscript{2}, Liansheng Wang\textsuperscript{1}\sthanks{Correspondence: lswang@xmu.edu.cn}, Yefeng Zheng\textsuperscript{2}}
\address{\textsuperscript{1}Xiamen University; \textsuperscript{2}Tencent Jarvis Lab; \textsuperscript{3}Jilin University}
\let\OLDthebibliography\thebibliography
\renewcommand\thebibliography[1]{
  \OLDthebibliography{#1}
  \setlength{\parskip}{0pt}
  \setlength{\itemsep}{0pt plus 0.3ex}
}

\begin{document}

\setlength{\abovedisplayskip}{0pt}
\setlength{\belowdisplayskip}{0pt}
\setlength{\abovedisplayshortskip}{0pt}
\setlength{\belowdisplayshortskip}{0pt}\noindent

%\ninept
%
\maketitle
\begin{abstract}
%Medical image segmentation plays an important role in clinical diagnosis, treatment planning, disease monitoring, \textit{etc}.
% Semantic segmentation plays an important role in medical image analysis.
Semantic segmentation is important in medical image analysis.
% Benefiting from recent advancements in deep learning (DL) techniques, fully convolutional networks (FCNs) have achieved compelling results in many segmentation tasks.
Inspired by the strong ability of traditional image analysis techniques in capturing shape priors and inter-subject similarity, many deep learning (DL) models have been recently proposed to exploit such prior information and achieved robust performance.
However, these two types of important prior information
%have not been explicitly captured and
are usually studied separately in existing models.
In this paper, we propose a novel DL model to model both type of priors within a single framework.
Specifically, we introduce an extra encoder into the classic encoder-decoder structure to form a Siamese structure for the encoders,
where one of them takes a target image as input (the image-encoder), and the other concatenates a template image and its foreground regions as input (the template-encoder).
%and besides the image branch (named the image-encoder) that takes a target image as input, we concatenate the foreground regions of a template image attached with itself to form an extra input to the extra branch (named the template-encoder).
The template-encoder encodes the shape priors and appearance characteristics 
% (such as intensity information)
 of each foreground class in the template image.
A cosine similarity based attention module is proposed to fuse the information from both encoders, to utilize both types of prior information encoded by the template-encoder and model the inter-subject similarity for each foreground class.
Extensive experiments on two public datasets demonstrate that our proposed method 
% is robust and 
can produce superior performance to competing methods.
\blfootnote{C. Xie and H. Liu contributed equally during internships at Tencent.}
\end{abstract}
\vspace{-1.2em}
\section{Introduction}
\vspace{-0.5em}
Semantic segmentation, which is aimed at predicting semantic labels for pixels within images, is a fundamental problem in medical image analysis.
With the development of deep fully convolutional networks (FCN \cite{cciccek20163d}), compelling results have been achieved by introducing contextual information into deep learning (DL) for image segmentation.
% Despite great advances in medical image
% segmentation, many challenges still remain which recent works \cite{bentaieb2016topology,ravishankar2017learning,ambellan2019automated,balsiger2019learning,dinsdale2019spatial,yang2018neural,li2019hybrid,wang2019pairwise} have tried to address.
Despite great advances in medical image
segmentation, many challenges still remain which recent works have tried to address.
First, shape priors are not effectively captured.
Second, inter-subject similarity, which provided abundant resources for anatomical priors in classical medical image segmentation methods such as atlas-based segmentation 
% \cite{yang2018neural,liang2019comparenet}
\cite{yang2018neural}
, is not fully-exploited.
Third, DL methods usually rely on a large amount of representative data for training.
%to intrinsically model semantics of segmentation tasks.
With limited training data, it is difficult to balance between exactly modeling the prior knowledge of target objects and robustly representing individual differences.

Inspired by traditional segmentation techniques that incorporated shape and/or appearance prior knowledge, many DL models were proposed to exploit such prior knowledge for improving accuracy and robustness.
One group of methods modeled inter-subject similarity by fully exploiting anatomical priors.
These works were motivated by the classical concept of atlas-based segmentation, and implemented in two ways: formulating atlas-based segmentation with a DL framework %\cite{yang2018neural,vakalopoulou2018atlasnet,liang2019comparenet,duan2019automatic},
% \cite{yang2018neural,liang2019comparenet},
\cite{yang2018neural},
and combining the segmentation task with a registration task for joint optimization %\cite{li2019hybrid,estienne2019u,xu2019deepatlas,elmahdy2019adversarial}
% \cite{li2019hybrid,xu2019deepatlas}.
\cite{xu2019deepatlas}.
Another group of methods leveraged shape priors from different perspectives, \textit{e.g.,} formulating a new loss function
%\cite{bentaieb2016topology,ravishankar2017joint,yang2019right}
% \cite{bentaieb2016topology,ravishankar2017joint},
\cite{ravishankar2017joint},
imposing constraints on shape codes/representations in latent space
%\cite{ravishankar2017learning,oktay2017anatomically,he20183d,yue2019cardiac,chen2019learning},
% \cite{ravishankar2017learning,oktay2017anatomically}, 
\cite{ravishankar2017learning}, 
combining with statistical shape models~\cite{ambellan2019automated}, representing sparse anatomical shapes with point clouds~\cite{balsiger2019learning}, and learning a deformation field to resample an initial binary mask \cite{dinsdale2019spatial}.
A common drawback of all these methods was that the shape priors and inter-subject similarity were not simultaneously modeled.

\begin{figure}[!t]
\center
\includegraphics[width=1.0\columnwidth]{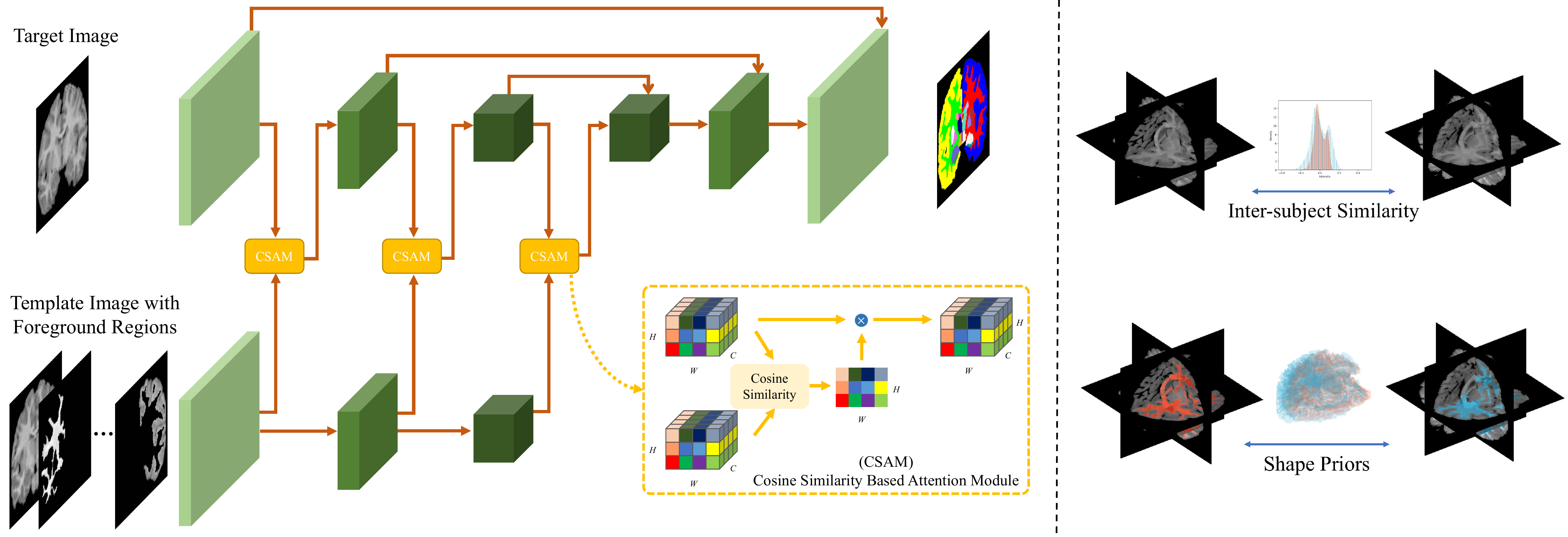}
\vspace{-1em}
\caption{Left: overview of the proposed Prior-Net for encoding shape priors and inter-subject similarity for segmentation.
Note that all operations are 3D, while being illustrated in 2D for simplicity.
Right: illustration of the inter-subject similarity and shape priors. Top: data from two different subjects are similar to each other in appearance and intensity distribution (normalized to zero mean and unit standard deviation); bottom: data from two different subjects share similar anatomical structures (demonstrated with right cerebral white matter).}
\label{fig_framework}
\vspace{-1.8em}
\end{figure}

So far, we are aware of only one work \cite{wang2019pairwise} closely related to ours, where the shape priors and inter-subject similarity were both leveraged within a single DL framework.
Specifically, a conjugate fully convolutional network (CFCN, a Siamese structure) \cite{wang2019pairwise} was proposed by Wang \textit{et al.}, which took pairs of samples as input to capture the inter-subject similarity, and a proxy supervision signal was formulated to encode the shape priors in the label space.
The limitations of this work are threefold.
First, the inter-subject similarity was modeled between slices, rather than volumes.
Hence, the 3D spatial consistency was not fully attended, resulting in compromised segmentation accuracy.
Second, this method only addressed binary segmentation, and it may be difficult to generalize for multi-class segmentation since the competition between classes and the class-imbalance may make the training process hard to converge.
%and it seemed that generalizing to multi-class would make the training hard to converge due to the competition between classes and the class-imbalance problem.
Third, the shape priors were only encoded in the label space, and the appearance characteristics of any specific tissue/organ were overlooked.

In this paper, we similarly adopt a Siamese structure as \cite{wang2019pairwise}---but only for the encoders---to encode shape priors and the inter-subject similarity, with a more effective implementation.
Specifically, one of the Siamese encoders takes a target image as input (the image-encoder), and the other concatenates a template image and its foreground regions as input (the template-encoder).
%besides the image branch (named the image-encoder) that takes a target image as input, we concatenate the foreground regions of each class of a template image attached with itself to form an extra input to another branch (named the template-encoder).
The concatenated input to the template-encoder incorporates the foreground regions of each class in the template image, enabling us to not only encode shape priors explicitly (thus eliminating the need for the proxy supervision as in \cite{wang2019pairwise}), but also exploit the appearance characteristics of each foreground region by the template-encoder.
In this sense, we name our network as Prior-Net.
In addition, a cosine similarity based attention module is proposed to fuse the information from both encoders, for fully exploiting the shape priors and appearance characteristics from the template-encoder, and modeling the inter-subject similarity between subjects for each foreground class.
We demonstrate the effectiveness of our proposed approach
%compared with other recent methods
on the publicly available Child and Adolescent NeuroDevelopment Initiative (CANDI) dataset \cite{kennedy2011candishare} and Liver Tumor Segmentation Challenge (LiTS) dataset \cite{bilic2019liver}, and experimentally show that our approach is robust, and capable of achieving superior performance to other methods.

\vspace{-0.7em}
\section{Method}
\vspace{-0.5em}
In this section, we present the details of the proposed Prior-Net. An overview of Prior-Net is shown in Fig. \ref{fig_framework} left.
The Prior-Net focuses on two types of important prior knowledge: shape priors and inter-subject similarity (see Fig. \ref{fig_framework} right for a visual illustration), and consists of three parts: an encoder-decoder structure (the image-encoder and image-decoder), an extra template-encoder, and a cosine similarity based attention module.
The encoder-decoder structure takes a sample as input, and extracts features with bottom-up and top-down ways for predicting the segmentation for each voxel of the input.
The extra template-encoder takes a template image and its foreground regions (obtained from the segmentation ground truth)
%foreground regions of a template image (obtained by the segmentation ground truth) attached with itself
as input to encode shape priors and appearance characteristics.
The cosine similarity based attention module is used to fuse the information from both encoders.
% branches for fully-exploiting the shape priors and appearance characteristics from the template-encoder and modeling the inter-subject similarity between subjects for each foreground class.

\noindent\textbf{Basic Encoder-Decoder Structure:}
We adopt the 3D U-Net \cite{cciccek20163d} as the basic encoder-decoder structure to learn deep semantics from the image and output segmentation.
%{\color{red}As known, the 3D U-Net \cite{cciccek20163d} is extended from the classic 2D U-Net \cite{ronneberger2015u} by replacing all 2D operations with 3D counterparts, and employing batch normalization \cite{ioffe2015batch} for faster convergence.}
The U-Net architecture contains a contracting path (\textit{i.e.,} the encoder), an expanding path (\textit{i.e.,} the decoder), and skip connections from the contracting to the expanding path at the same scale.
%The U-Net architecture contains a contracting path (\textit{i.e.,} encoder) with convolutions and downsampling, an expanding path (\textit{i.e.,} decoder) with convolutions and upsampling, and skip connections from the contracting path to the expanding path of the same scale.
In our experiments, to relieve the overfitting problem and reduce computation cost, we only use an overall downsampling rate of eight and half of the channel numbers in all layers compared to original implementation in \cite{cciccek20163d}.

\noindent\textbf{Extra Template-Encoder:}
As aforementioned, shape priors and appearance characteristics are important to the segmentation task. Different from \cite{wang2019pairwise}, which encodes the shape priors by proxy supervision on the label space, our work extensively exploits the appearance characteristics in the image space.
Utilizing the ground truth segmentation, we extract regions of each foreground class to output a separate image for each class, in which only regions containing the specific class are preserved with others set to zero.
%Similar to \cite{wang2019pairwise},
An extra encoder branch (named template-encoder)
%but without decoder
is employed to encodes shape priors.
Concretely, the template-encoder takes the foreground regions of each class of a template image attached with itself (via concatenation) as the input, and learns the shape priors and appearance characteristics from the template image, which are further fused with the image encoder to provide strong priors for learning the segmentation of the current image.
Note that, we also input the original template image to preserve the spatial correlations between classes.
In the implementation, we adopt the same network structure as the image encoder for the template encoder.

\noindent\textbf{Cosine Similarity Based Attention Module:}
Our work borrows ideas from the classical atlas-based segmentation \cite{iglesias2015multi}, which becomes once more prosperous 
% \cite{liang2019comparenet,yang2018neural}
\cite{yang2018neural}
 with the development of DL. Different from atlas-based segmentation, which usually obtains the segmentation result via indirectly learned correspondences between two images, we directly model the inter-subject similarity in the feature space.

Our aim is to explore strategies to fuse the high-level shape priors distilled from the template-encoder and model the inter-subject similarity to provide guidance for the segmentation of the target image.
Cosine similarity is usually used in the feature space to highlight the correlations between two features.
Actually, 
% it is widely used in deep pair-based metric learning \cite{wang2019multi} and deep image hashing \cite{gattupalli2019weakly}
it is widely used in deep image hashing \cite{gattupalli2019weakly}
, with the assumption that cosine similarity between the vectors of objects should be in accordance with their semantic similarity. Inspired by this, we propose a cosine similarity based attention module (CSAM) to fuse information from different branches. Specifically, CSAM is implemented with an attention on the features in the image-encoder, where the attention weights are computed by the cosine similarity between features in the same positions of the two feature maps from both encoders.
Formally, given the feature maps $f^1, f^2 \in \mathbb{R}^{H \times W \times D \times C}$ in the image-encoder and template-encoder, respectively, the attention weights $w \in \mathbb{R}^{H \times W \times D}$ can be computed by
%Formally, given a feature map $f^1 \in \mathbb{R}^{H \times W \times D \times C}$ in the image-encoder and a feature map $f^2 \in \mathbb{R}^{H \times W \times D \times C}$ in the template-encoder, the attention weights $w \in \mathbb{R}^{H \times W \times D}$ can be computed by
%\begin{equation}
%w_{i, j, k}=\frac{f^1_{i,j,k} \cdot f^2_{i,j,k}}{\left\| f^1_{i,j,k}\right\|_2 \cdot \left\|f^2_{i,j,k}\right\|_2},
%\end{equation}
\begin{equation}
w_{i, j, k}={\big(f^1_{i,j,k} \cdot f^2_{i,j,k}\big)}\Big/{\big(\big\| f^1_{i,j,k}\big\|_2 \cdot \big\|f^2_{i,j,k}\big\|_2\big)},
\end{equation}
where $H,W,D$ are the height, width, and depth of the feature maps, respectively, with $i,j,k$ as indices, and $C$ is the channel number.
As shown in Fig. \ref{fig_framework}, we employ CSAM in all levels of both encoders.
Such a design enables us to encode the abstract prior knowledge in a hierarchical way, and as the network level goes deeper, the pairwise similarity would be more accurate with deeper semantics.

\noindent\textbf{Training and Inference Details:}
In the training phase, the target and template images are randomly sampled from the training set, and trained with a multi-class Dice loss \cite{isensee2018nnu}.
In the test phase, one image in the training set is randomly sampled as the template image, and coupled with a target image in the test set as input to Prior-Net for segmenting the target image.
Note that we also experiment with different template images for prediction, and the results show negligible variations, demonstrating the robustness of Prior-Net.
% Thus, label fusion strategies in multi-atlas segmentation \cite{yang2018neural,vakalopoulou2018atlasnet,iglesias2015multi} can be also explored in our Prior-Net. In the experiment parts, we will present results with randomly sampled template images to demonstrate this in situations with limited data.

\vspace{-0.5em}
\section{Experiments}
\vspace{-0.5em}
\noindent\textbf{Datasets:}
\textit{CANDI}~\cite{kennedy2011candishare} comprises 103 T1-weighted MRI scans with anatomic segmentation labels.
% \textit{CANDI}~\cite{kennedy2011candishare} is a public dataset from
% the Child and Adolescent NeuroDevelopment Initiative at the University of Massachusetts Medical School, comprising 103 T1-weighted MRI scans (57 males and 46 females) with anatomic segmentation labels.
% The subjects come from four diagnostic groups: healthy controls, schizophrenia spectrum, bipolar disorder with psychosis, and bipolar disorder without psychosis.
We use 28 anatomical structures that were used in VoxelMorph \cite{balakrishnan2019voxelmorph}. 
% The volume size ranges from $256 \times 256 \times 128$ to $256 \times 256 \times 158$ voxels.
For computational efficiency, we crop the central region of each scan with a size of $160 \times 160 \times 128$ voxels, which is large enough to contain the whole brain.
We randomly select 21 volumes as test data, and use the others for training.
\textit{LiTS}~\cite{bilic2019liver} comprises 201 contrast-enhanced abdominal CT scans.
% \textit{LiTS}~\cite{bilic2019liver} is a publicly available liver tumor dataset comprising 201 contrast-enhanced abdominal CT scans collected from various clinical sites around the world. 
The dataset was originally split into a training set (131 scans) and a testing set (70 scans), where only the training set was publicly available with accurate liver and lesion masks.
For simplicity, we conduct experiments only on liver regions to verify the effectiveness of our method. We randomly select 26 volumes as test data, and use the others for training as CFCN \cite{wang2019pairwise}.

% \subsubsection{Evaluation Metric.}

%\subsection{Implementation Details}
\noindent\textbf{Implementation:}
For the LiTS dataset \cite{bilic2019liver}, we truncate the Hounsfield unit of all scans outside the range of [$-$200, 250] to ignore irrelevant content.
For the CANDI dataset \cite{kennedy2011candishare}, we normalize the volumes to zero mean and unit standard deviation.
% The Dice coefficient score is used as the evaluation metric.
All experiments are implemented with PyTorch framework.
We use the Adam optimizer \cite{kingma2014adam} to train our model for 60 epochs. The batch size is two with an NVIDIA Tesla M40 GPU. The cosine annealing technique is adapted to adjust the learning rate from 0.02 to $1 \times 10^{-6}$.

\noindent\textbf{Ablation Study on Prior Knowledge Encoding:}
We conduct an ablation study to examine the effectiveness of Prior-Net in modeling prior knowledge on the CANDI dataset, and identify the contribution of each building block.
We employ 3D U-Net \cite{cciccek20163d} as the no-prior baseline.
Since we propose to exploit prior knowledge from a template image with foreground regions of each class, an immediate improvement is to concatenate them with the target image as input to the 3D U-Net for training (single encoder).
Next, we extend the 3D-Net with an extra template-encoder to separately learn high-level shape priors, where the deep semantics from the template-encoder are simply fused into the image-encoder by concatenation (dual encoder).
Lastly, inspired by the wide adoption of cosine similarity in feature matching 
% \cite{wang2019multi,gattupalli2019weakly}
\cite{gattupalli2019weakly}
, we propose the CSAM to extract abstract prior knowledge with pairwise similarity guidance in a hierarchical way (the proposed Prior-Net).
We show the results in Table \ref{exp:ablation_study}. From the table, we can observe that even with only a single encoder, our prior modeling can bring a 2.38\% improvement in average Dice score.
With the dual encoder and CSAMs, further improvements of 1.10\% and 2.49\% are achieved in average Dice score, respectively.
The consistent performance improvements demonstrate the significance of Prior-Net and its CSAM module in modeling prior knowledge.

\begin{table}[!t]\vspace{-2.5mm}
  \centering
  \caption{Ablation study on prior knowledge encoding on the CANDI dataset \cite{kennedy2011candishare}.}\label{exp:ablation_study}
  \setlength{\tabcolsep}{.7mm}
  \scalebox{0.9}{\centering
  \begin{tabular}{ccccc}
    \hline
    Method & Baseline \cite{cciccek20163d} & Single encoder & Dual encoder & Prior-Net \\
    & (no prior) & (variant of ours) & (variant of ours) & (ours) \\
    \hline
    Dice (\%) & 83.10 & 85.48 & 86.58 & 89.07 \\
    \hline
  \end{tabular}}
  \vspace{-1em}
\end{table}

\begin{figure}[!t]
\center
\includegraphics[width=.95\columnwidth]{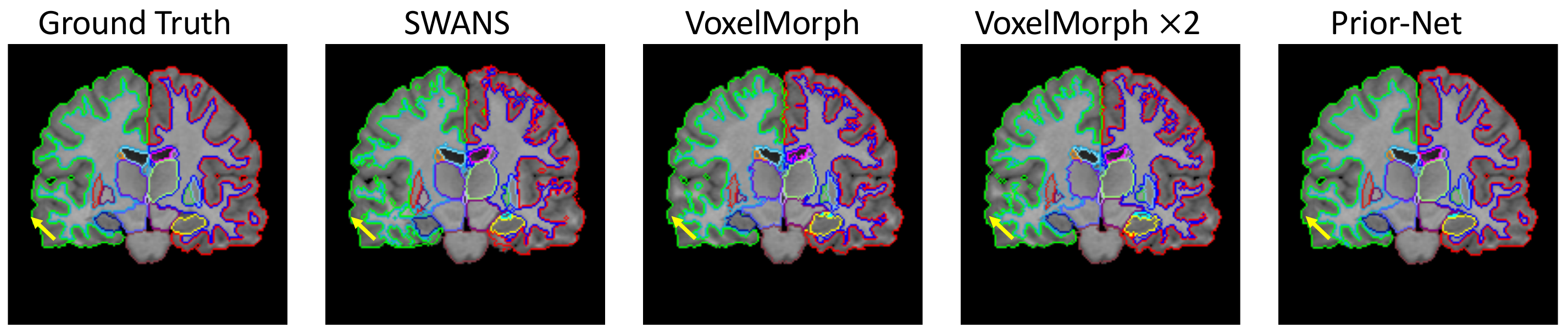}
\vspace{-0.5em}
\caption{Example MR slice of brain structure annotations and segmentation maps predicted by SWANS \cite{dinsdale2019spatial}, VoxelMorph \cite{balakrishnan2019voxelmorph}, VoxelMorph $\times 2$ \cite{balakrishnan2019voxelmorph} and our Prior-Net. 
Yellow arrows point to flaws in the predictions made by other methods, as compared to those by Prior-Net.}
\label{fig_results}
\vspace{-1.8em}
\end{figure}

\noindent\textbf{Comparison with Other Methods:}
We compare with two state-of-the-art methods: SWANS \cite{dinsdale2019spatial} which models shape priors, and VoxelMorph \cite{balakrishnan2019voxelmorph} which models inter-subject similarity, on the CANDI dataset.
Although VoxelMorph was mainly focused on registration, it also presented solutions to using ground truth segmentation for training.
To adapt VoxelMorph for atlas-based segmentation, we exchange the positions of the moving and fixed images (corresponding to the target and template images in our context), and use the learned registration field to warp the segmentation of the fixed image to get the segmentation of the moving (target) image.
In contrast, Prior-Net encodes both types of prior knowledge within a single DL framework.
The results are shown in Table \ref{exp:candi_sota}. We also present results with a variant of VoxelMorph (VoxelMorph $\times 2$), which doubled the number of features to account for the increased inherent variability of the task \cite{balakrishnan2019voxelmorph}.
From the table, we can observe that our Prior-Net substantially boosts the performance, \textit{e.g.,} with a 6.67\% improvement in average Dice score over SWANS \cite{dinsdale2019spatial}.
The reasons may be that: 1) SWANS was originally proposed for binary segmentation, and might fail to generalize to multi-class segmentation due to competition of different classes in learning the deformation field;
2) VoxelMorph was proposed for registration, where the segmentation learning is not well-supervised.
We visualize some example slices of brain structure annotations and segmentation maps predicted by the different methods in Fig. \ref{fig_results}.
Compared to the competing methods, Prior-Net predicts brain structures that are more anatomically meaningful.

\begin{table}[!t]\vspace{-2.5mm}
  \centering
  \caption{Comparing Prior-Net with SWANS \cite{dinsdale2019spatial} and VoxelMorph \cite{balakrishnan2019voxelmorph} on the CANDI dataset \cite{kennedy2011candishare}.}\label{exp:candi_sota}
  \setlength{\tabcolsep}{.77mm}
  \scalebox{0.9}{\centering
  \begin{tabular}{ccccc}
    \hline
    Method & SWANS & VoxelMorph &  VoxelMorph $\times 2$ & Prior-Net (ours) \\
    \hline
    Dice (\%) & 82.40 & 79.96 & 81.13 & 89.07 \\
    \hline
  \end{tabular}}
  \vspace{-1.2em}
\end{table}

\noindent\textbf{Comparison with CFCN on the LiTS Dataset:}
Our work is partially motivated by CFCN, in which shape priors and inter-subject similarity were explored with a similar Siamese structure and an extra proxy supervision in the label space.
For a fair comparison to CFCN, we adapt our Prior-Net for 2D images, and employ the same sampling strategy to form a template and target image pair.
As done in \cite{wang2019pairwise}, we present results with different ratios of data for training, \textit{i.e.,} 1\%, 5\%, and 80\% of
the official training subset (corresponding to 1, 7, 105 volumes, respectively).
From Table \ref{tab_compare_cfcn}, we can observe that Prior-Net can also achieve reasonable segmentation accuracies with a limited number of training samples, which are better than CFCN.
This is expected, since the fusion strategies of Prior-Net are explored in the encoder parts, and with CSAMs, the shape priors and inter-subject similarity are gradually fused into the image encoder from the template image in a hierarchical way;
whereas for CFCN, the fusion strategies were implemented in the decoder and loss function, making it difficult for the encoder to fit, and small turbulence might dramatically affect the encoder.
Besides, foreground regions of each class are explicitly modeled in our work, which may also benefit the performance.

\begin{table}[!t]\vspace{-2mm}
  \centering
  \caption{Comparing Prior-Net with CFCN \cite{wang2019pairwise} on the LiTS dataset with 1\%, 5\%, and 80\% of official training samples (Dice (\%)).
  We use a different random split of the training and test sets from \cite{wang2019pairwise};
  therefore, the results reported here present minor differences from the original numbers in \cite{wang2019pairwise}.} \label{tab_compare_cfcn}
  \scalebox{0.9}{\centering
  \begin{tabular}{cccc}
    \hline
    Training sample ratio & 1\% & 5\% & 80\% \\
    \hline
    CFCN \cite{wang2019pairwise} & 88.40 & 94.12 & 95.33 \\
    Prior-Net (2D) & 90.24 & 95.06 & 96.48 \\
    \hline
  \end{tabular}}\vspace{-1.5em}
  \end{table}  

\vspace{-0.8em}
\section{Conclusion}
\vspace{-0.5em}
This paper proposed to model shape priors and inter-subject similarity with Prior-Net to overcome the limitation of DL with a limited number of annotated data.
Extensive experiments on two publicly available datasets demonstrated the effectiveness of Prior-Net in modeling prior knowledge.

\vspace{-0.8em}
\section{Compliance with Ethical Standards}
\vspace{-0.8em}
This research study was conducted retrospectively using human subject data made available in open access by \cite{kennedy2011candishare,bilic2019liver}.
Ethical approval was not required as confirmed by the license attached with the open access data.

\noindent\textbf{Acknowledgments.}
This work was supported by the Fundamental Research Funds for the Central Universities (Grant No. 20720190012).
Shilei Cao, Dong Wei, Kai Ma and Yefeng Zheng are employees of Tencent.

\vspace{-0.5em}
\bibliographystyle{IEEEbib}
\bibliography{strings,mybib}

\end{document}